\documentclass[10pt,twocolumn,letterpaper]{article}

\usepackage{cvpr}              

\usepackage[utf8]{inputenc} 
\usepackage[T1]{fontenc}    
\usepackage{url}            
\usepackage{booktabs}       
\usepackage{amsfonts}       
\usepackage{nicefrac}       
\usepackage{microtype}      
\usepackage{xcolor}         
\usepackage{epstopdf}
\usepackage{multirow}
\usepackage{colortbl}
\usepackage[pagebackref=true,breaklinks=true,letterpaper=true,colorlinks,bookmarks=false]{hyperref}
\usepackage{subcaption}
\usepackage{amsmath} 
\usepackage{amssymb}            
\usepackage{mathrsfs}            
\usepackage{bm}
\usepackage{wrapfig}
\usepackage{bbm}
\usepackage[capitalize]{cleveref}
\usepackage{algorithm}
\usepackage{algorithmic}
\usepackage{marvosym}
\usepackage{inconsolata}

\makeatletter
\newcommand*\bigcdot{\mathpalette\bigcdot@{.5}}
\newcommand*\bigcdot@[2]{\mathbin{\vcenter{\hbox{\scalebox{#2}{$\m@th#1\bullet$}}}}}
\makeatother

\definecolor{c2}{HTML}{FBD9BD}
\definecolor{c3}{HTML}{fe793d}
\definecolor{c4}{HTML}{eedeb0}
\definecolor{pp}{HTML}{BC7FCD}
\definecolor{bb}{HTML}{CDE8E5}
\definecolor{c5}{HTML}{00FFFF}
\definecolor{c6}{HTML}{FF00FF}
\definecolor{rouse}{rgb}{0.981,0.961,0.941}

\crefname{section}{Sec.}{Secs.}
\Crefname{section}{Section}{Sections}
\Crefname{table}{Tab.}{Tabs.}
\crefname{table}{Table}{Tables}
\crefname{figure}{Fig.}{Figs.}
\crefname{equation}{Eq.}{Eqs.}

\definecolor{cvprblue}{rgb}{0.21,0.49,0.74}

\def\paperID{} 
\def\confName{CVPR}
\def\confYear{2026}

\title{SCALER: SAM-Enhanced Collaborative Learning for \\ Label-Deficient Concealed Object Segmentation}

\author{Chunming He$^{1}$\,,
        Rihan Zhang$^{1}$\,,
        Longxiang Tang$^{2}$\,,
        Ziyun Yang$^{3}$\,,  \\
	{Kai Li}$^{4}$\,, 
            {Deng-Ping Fan}$^{5}$ \,, 
        and {Sina Farsiu}$^{1,\dagger}$\\
        $^1$Duke University,
	$^2$Harvard University,  
        $^3$Apple, 
        $^4$Meta, 
        $^5$Nankai University, \\
$\dagger$ Corresponding Author, Contact: chunming.he@duke.edu.
}

\begin{document}
\twocolumn[{
\maketitle
\vspace{-12mm}
\begin{center}
\includegraphics[width=1\linewidth]{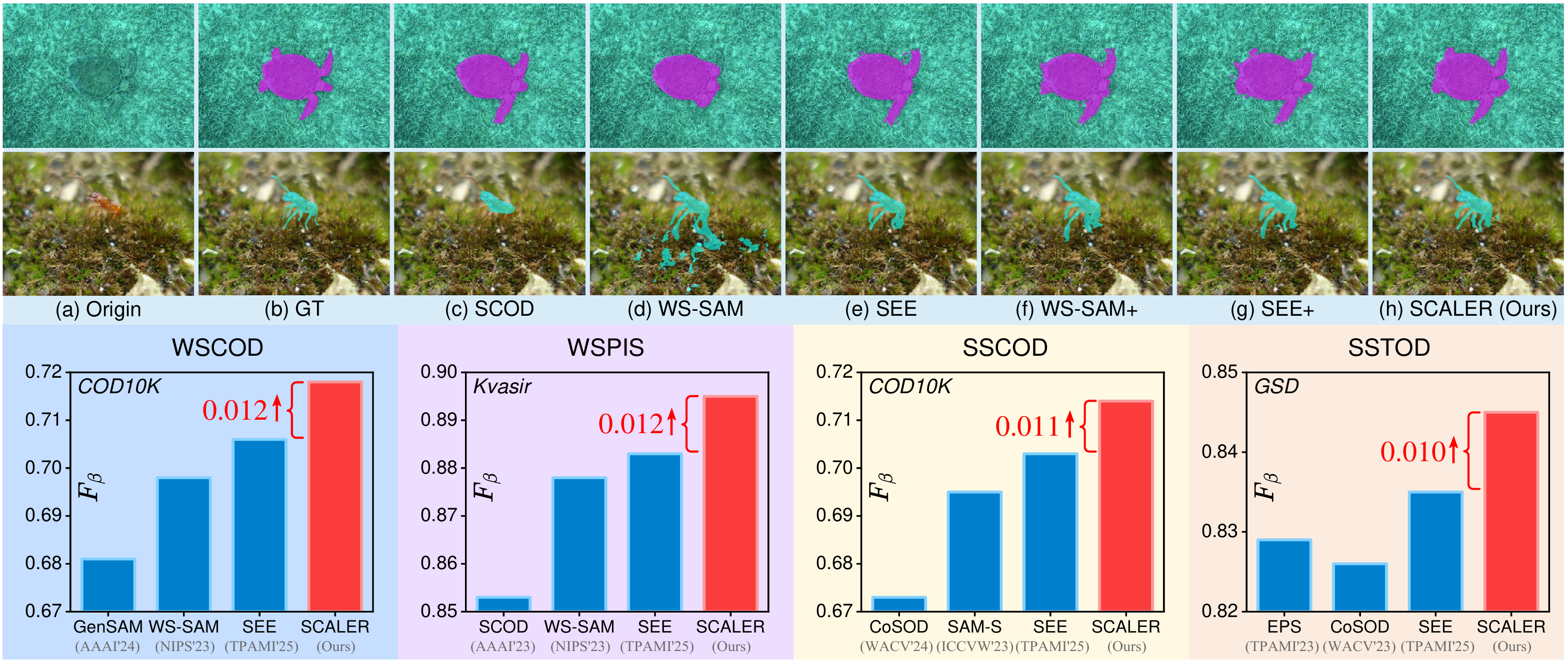}
\vspace{-8mm}    
	\captionof{figure}{Results of existing LDCOS methods with point supervision, including SCOD~\cite{he2022weakly}, WS-SAM~\cite{he2023weaklysupervised}, and SEE~\cite{he2025segment}. The suffix ``+'' denotes integration with SCALER. SCALER yields more accurate concealed‑object segmentation and achieves leading results across COD (camouflaged object detection), PIS (polyp image segmentation), and TOD (transparent object detection) under weak (WS) and semi‑supervised (SS) settings. In the top section, concealed objects masks are highlighted in {\textcolor{c6}{pink}} and {\textcolor{c5}{blue}}.}
    \label{fig:Intro}
\end{center}
}]

\begin{abstract} \label{abstract}
Existing methods for label-deficient concealed object segmentation (LDCOS) either rely on consistency constraints or Segment Anything Model (SAM)-based pseudo-labeling. However, their performance remains limited due to the intrinsic concealment of targets and the scarcity of annotations. This study investigates two key questions: 
(1) Can consistency constraints and SAM-based supervision be jointly integrated to better exploit complementary information and enhance the segmenter? and (2) beyond that, can the segmenter in turn guide SAM through reciprocal supervision, enabling mutual improvement?
To answer these questions, we present SCALER, a unified collaborative framework toward LDCOS that jointly optimizes a mean-teacher segmenter and a learnable SAM. SCALER operates in two alternating phases. In \textbf{Phase \uppercase\expandafter{\romannumeral1}}, the segmenter is optimized under fixed SAM supervision using entropy-based image-level and uncertainty-based pixel-level weighting to select reliable pseudo-label regions and emphasize harder examples. In \textbf{Phase \uppercase\expandafter{\romannumeral2}}, SAM is updated via augmentation invariance and noise resistance losses, leveraging its inherent robustness to perturbations.
Experiments demonstrate that SCALER yields consistent performance gains across eight semi- and weakly-supervised COS tasks. The results further suggest that SCALER can serve as a general training paradigm to enhance both lightweight segmenters and large foundation models under label-scarce conditions. Code will be released.

\end{abstract}
\vspace{-4mm}
\setlength{\abovedisplayskip}{2pt}
\setlength{\belowdisplayskip}{2pt}
\section{Introduction} \label{introduction}
Concealed object segmentation (COS) encompasses scenarios where targets share high visual similarity with their surroundings, making them difficult to distinguish. Representative tasks include camouflaged object detection (COD) \cite{fan2020camouflaged,He2023Camouflaged}, polyp image segmentation (PIS) \cite{fan2020pranet,xiao2023concealed}, medical tubular object segmentation (MTOS) \cite{he2023HQG,he2025run}, and transparent object detection (TOD) \cite{he2023weaklysupervised,xiao2024survey}. 
Unlike conventional segmentation, COS must contend with weak contrast and ambiguous boundaries, which significantly complicate training. Early studies mitigated these difficulties through perceptual priors \cite{he2023strategic,zhao2025focusdiffuser} and auxiliary cues \cite{He2023Camouflaged,sun2025frequency}, showing strong results under full supervision.

The challenge grows sharper in label-deficient COS (LDCOS), including semi- (SSCOS) and weakly-supervised (WSCOS) setups, where limited annotations constrain learning. Prior works have extended consistency-based frameworks, particularly the mean-teacher paradigm that enforces prediction stability between a student and an exponentially averaged teacher~\cite{he2022weakly,lai2024camoteacher}. However, the reliance on pseudo-labels often introduces error accumulation, especially for concealed targets with uncertain boundaries (see~\cref{fig:Intro}).

\begin{figure}[t]
\centering
\setlength{\abovecaptionskip}{0cm} 	\setlength{\belowcaptionskip}{0cm}
\includegraphics[width=0.8\linewidth]{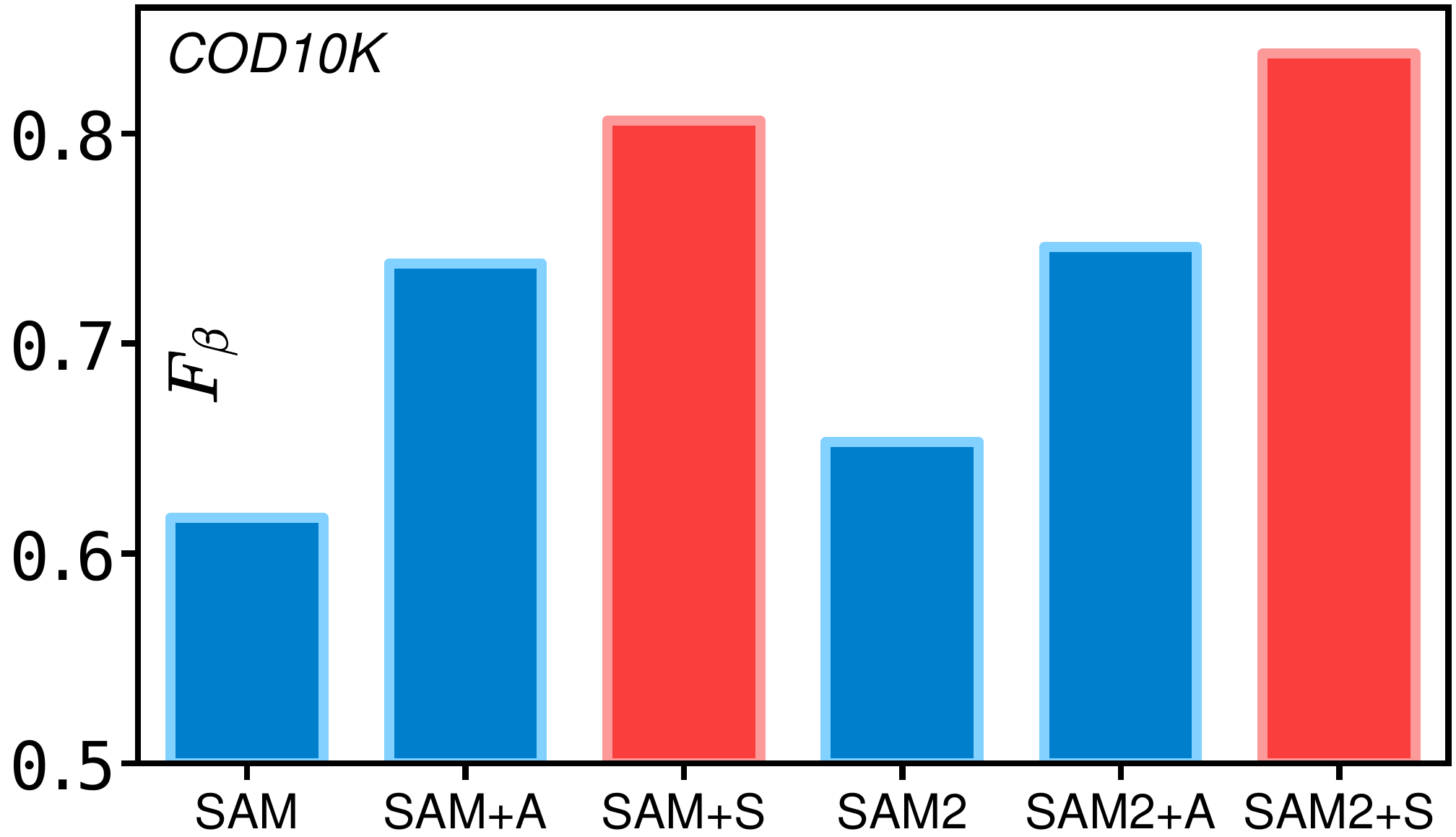}
	\caption{Quality of pseudo-labels from different SAM variants. 
    }
    \label{fig:QualiPL}
	\vspace{-4mm}
\end{figure}

To exploit foundation model guidance, recent works such as WS-SAM\cite{he2023weaklysupervised} and SEE\cite{he2025segment} employed the pretrained Segment Anything Model (SAM)\cite{kirillov2023segment} as an external teacher, transferring its global knowledge to task-specific segmenters. These methods confirm SAM’s potential in providing supplementary supervision, yet face two major limitations (see \cref{fig:QualiPL}). First, SAM’s fixed predictions are not tailored for COS, yielding noisy and suboptimal pseudo labels. Second, the information flow remains one-way, from SAM to the segmenter, without allowing the evolving, task-adapted segmenter to reciprocally benefit SAM.

These shortcomings raise two central questions: \textit{(1) Can consistency learning and SAM-based supervision be jointly leveraged to improve segmenters under label scarcity?} \textit{(2) Can the segmenter itself provide pseudo-label feedback to strengthen SAM, enabling mutual improvement?}

To address these questions, we propose \textbf{SCALER}, a SAM-enhanced collaborative learning framework for LDCOS. SCALER integrates the consistency regularization of the mean-teacher model with a learnable SAM for knowledge distillation and reciprocal adaptation.
They are optimized alternately in two phases, with pseudo-label generation strategies tailored to their respective properties.

In \textbf{Phase \uppercase\expandafter{\romannumeral1}}, we train the segmenter within the mean-teacher framework while keeping SAM fixed. Given the segmenter’s sensitivity to label noise, we introduce an entropy-based image-level weighting mechanism and an uncertainty-based pixel-level weighting strategy to selectively leverage pseudo-labels generated by both the teacher network and the fixed SAM. These strategies aim to suppress unreliable regions and reduce the influence of low-quality supervision. Besides, for samples that are already well-segmented by the segmenter, we apply strong augmentations to mine potentially informative features previously overlooked.

In \textbf{Phase \uppercase\expandafter{\romannumeral2}}, we fix the mean-teacher framework and use it to generate pseudo-labels for updating SAM. To exploit SAM’s robustness to perturbations, we propose an augmentation invariance loss and a noise resistance loss. The augmentation invariance loss encourages SAM to maintain consistent predictions under weak and strong augmentations, facilitating robust representation learning without impairing generalization. The noise resistance loss enables SAM to utilize label-deficient data more effectively, even when pseudo-labels are noisy. By increasing the diversity and quantity of training samples, this strategy mitigates the overfitting that SAM may encounter under limited supervision.

By alternately optimizing the two phases, both the segmenter and SAM are progressively improved through mutual knowledge transfer. This iterative training scheme enables more effective utilization of label-deficient data, ultimately enhancing segmentation performance in COS tasks.
Our contributions can be summarized as follows:
\begin{itemize}[leftmargin=*]
  \item We propose a novel bi-directional collaborative learning framework, SCALER, that enables a downstream segmenter and a SAM-based model to achieve mutual enhancement in LDCOS tasks. We demonstrate this mechanism is effective for both SSCOS and WSCOS settings. 
  \item We design two specialized optimization phases that make this mutual learning possible. For the segmenter (Phase \uppercase\expandafter{\romannumeral1}), we introduce entropy- and uncertainty-based weighting strategies to filter noise from teacher pseudo-labels. For SAM (Phase \uppercase\expandafter{\romannumeral2}), we propose a novel augmentation invariance loss and noise resistance loss, allowing robust learning from the segmenter's noisy feedback.
  \item Extensive experiments on eight tasks validate our effectiveness and generalization, showing that our bi-directional SCALER outperforms prior one-way methods (WS-SAM and SEE) and successfully enhances the performance of vision foundation models like SAM and SAM2.
\end{itemize}

\begin{figure*}[t]
\setlength{\abovecaptionskip}{0cm} 	\setlength{\belowcaptionskip}{0cm}
\includegraphics[width=1\linewidth]{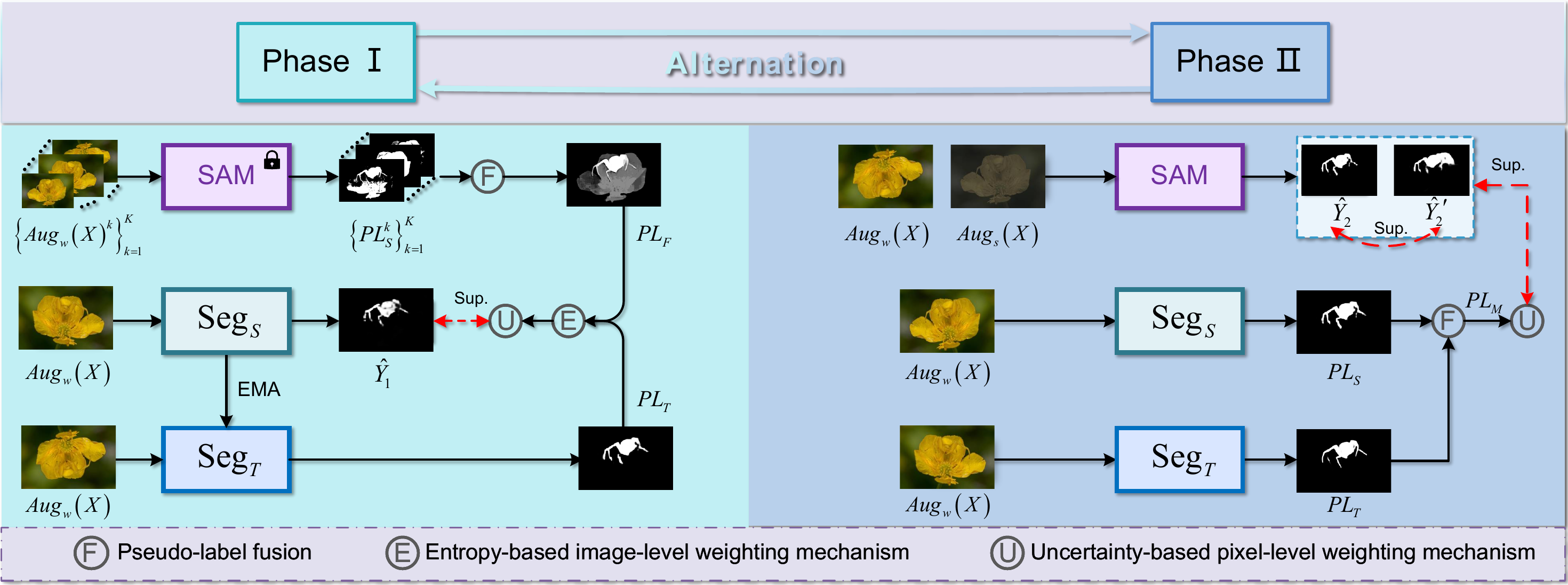}
	\captionof{figure}{The bi-directional collaborative learning framework of SCALER. The framework alternates between two optimization phases, which enables the models to mutually enhance each other.
    Notably, our framework supports the integration of various existing segmenters and SAM/SAM2 fine-tuning approaches. Augmentation strategies are randomly sampled to further enhance training flexibility. 
    }
    \label{fig:framework}
	\vspace{-4mm}
\end{figure*}

\vspace{-1mm}
\section{Related Work} \label{RelatedWork}
\noindent\textbf{Concealed Object Segmentation}. 
Learning-based segmenters have achieved notable progress in COS under full supervision \cite{fan2020camouflaged,He2023Camouflaged,he2023strategic,he2025run}.
Representative methods include SINet \cite{fan2020camouflaged} for reversible segmentation, FEDER \cite{He2023Camouflaged} for edge-assisted refinement, Camouflageator \cite{he2023strategic} for adversarial sample generation, and RUN \cite{he2025run} for model-driven optimization.
Recent LDCOS techniques, including semi- and weakly-supervised setups, mainly employ consistency regularization \cite{he2022weakly,lai2024camoteacher} or SAM-based pseudo-labeling~\cite{kirillov2023segment,he2023weaklysupervised,he2025segment,hu2024relax}.
However, unstable consistency and task-agnostic SAM supervision often lead to suboptimal segmentation.
To resolve this, SCALER unifies consistency regularization and adaptive SAM-based learning to achieve mutual refinement between the segmenter and the vision foundation model, promoting segmentation performance.

\noindent\textbf{Label-Deficient Image Segmentation}. Semi- and weakly-supervised segmentation improve label efficiency through regularization and pseudo-label refinement\cite{ren2023towards,basak2023pseudo,lee2023saliency,chakraborty2024unsupervised}. A common paradigm is the teacher–student family, especially the mean-teacher framework\cite{tarvainen2017mean}, where a teacher updated by an exponential moving average (EMA) of the student provides temporally ensembled targets for consistency learning. However, in COS, the ambiguous boundaries and low contrast destabilize pseudo-labels and amplify error accumulation. Meanwhile, fixed or lightly tuned foundation models such as SAM~\cite{kirillov2023segment} suffer domain gaps in concealed scenes. Recent methods, \textit{e.g.}, WS-SAM~\cite{he2023weaklysupervised} and SEE~\cite{he2025segment}, design complex pseudo-label filtering strategies to filter out valuable pseudo-labels generated by SAM, restricted by the limited capacity of SAM and suffering from limited training data. SCALER addresses these issues via alternating optimization: a mean-teacher module enforces robust consistency on the segmenter with uncertainty-aware weighting, while SAM is iteratively adapted with augmentation-invariance and noise-resistance losses. This collaborative design, unlike prior one-way distillation, enables bidirectional knowledge transfer and improves both specialized segmenters and foundation models in label-scarce scenarios.

\section{Methodology} \label{Sec:Methodology}
Label-deficient concealed object segmentation (LDCOS), including semi-supervised COS (SSCOS) and weakly supervised COS (WSCOS), aims to train a segmenter from a partially labeled training set $\mathcal{S}=\{X_i, Y_i\}_{i=1}^{S}$ and evaluate it on a test set $\mathcal{T}=\{T_i\}_{i=1}^{T}$, where $X_i$ and $T_i$ denote input images and $Y_i$ denotes the corresponding deficient label. In SSCOS, $Y_i$ is missing for some training samples, whereas in WSCOS, $Y_i$ comprises sparse annotations (\textit{e.g.}, points or scribbles) marking foreground and background regions.

Training an LDCOS model is challenging for two reasons. First, pure consistency constraints can be unstable due to the concealed nature of targets. Second, while SAM~\cite{kirillov2023segment} provides general priors, its predictions suffer from domain gaps. Prior methods~\cite{he2023weaklysupervised,he2025segment} address this through \textit{one-way} knowledge transfer (from SAM to the segmenter) but lack a mechanism for the segmenter, which becomes a domain specialist, to reciprocally improve SAM.

To overcome this, we propose \textbf{SCALER}, a \underline{S}AM-Enhanced \underline{C}oll\underline{A}borative \underline{LE}a\underline{R}ning framework that enables \textit{bi-directional} knowledge exchange. As illustrated in~\cref{fig:framework}, SCALER integrates a mean-teacher framework (for consistency) with a learnable SAM (for mutual distillation). The key innovation lies in alternating optimization across two phases: the segmenter learns from SAM (Phase~\uppercase\expandafter{\romannumeral1}) while SAM learns from the specialized segmenter (Phase~\uppercase\expandafter{\romannumeral2}), creating a mutually beneficial training loop.

\subsection{Phase \uppercase\expandafter{\romannumeral1}: Training Segmenter with Fixed SAM}
In Phase~\uppercase\expandafter{\romannumeral1}, we fix SAM (as a static ``generalist'' teacher) and optimize the segmenter. This segmenter benefits from dual-source supervision: (1) consistency regularization from its own teacher model ($\text{Seg}_T$) and (2) general knowledge from the frozen SAM (can be any vision foundation model).

\noindent\textbf{Dual pseudo-label generation}.
We generate two primary pseudo-labels.
First, for the teacher model $\text{Seg}_T$, we apply a random weak augmentation $Aug_w(\bigcdot)$ to each input ${X}$ and obtain the initial pseudo-label $PL_T$, formulated as:
\begin{equation}
    PL_T=\text{Seg}_T\left(Aug_w({X})\right).
\end{equation}
Next, for $\text{SAM}$, we generate $K$ stochastic augmentations of ${X}_i$, $\{Aug_w({X})^k\}_{k=1}^K$, by randomly sampling flips, rotations ($0^{\circ}$, $90^{\circ}$, $180^{\circ}$, $270^{\circ}$), and scales ($\times0.5$, $\times1.0$, $\times2.0$). Feeding these variants into SAM yields pseudo-labels:
\begin{equation}
    \{PL_S^k\}_{k=1}^K=\text{SAM}(\{Aug_w({X})^k\}_{k=1}^K).
\end{equation}
We then aggregate these masks by averaging:
\begin{equation}
    PL_F=\frac{1}{K}\sum_{k=1}^K PL_S^k.
\end{equation}
This ensemble strategy leverages SAM’s sensitivity to subtle variations~\cite{he2025segment}, yielding a $PL_F$ that is generally more accurate than any single prediction.

\noindent \textbf{Pseudo-label refinement}. Since initial pseudo-labels may be unreliable for some samples and pixel-level predictions, we propose two complementary weighting mechanisms: entropy-based image-level weighting and uncertainty-based pixel-level weighting. These mechanisms emphasize reliable regions within relatively easy-to-segment samples. Both strategies are applied to pseudo-labels generated by the teacher model and SAM, formulated as:
\begin{equation}\hspace{-3mm}
\scalebox{0.87}{$
\displaystyle
\begin{aligned}
    E(PL_T)\!=\!\frac{1}{N}\!\sum_i^N \!-\!PL_T^i\log_2 PL_T^i  -\!(1\!-\!PL_T^i)\log_2(1\!-\!PL_T^i\!),
	\end{aligned}
    $}
\end{equation}
\vspace{-2mm}
\begin{equation}\hspace{-3mm}
\scalebox{0.87}{$
\displaystyle
\begin{aligned}
    E(PL_F) \!=\! \frac{1}{N}\!\sum_i^N \!-\!PL_F^i\log_2 PL_F^i-\!(1\!-\!PL_F^i\!)\log_2(1\!-\!PL_F^i\!),
    \end{aligned}
    $}
\end{equation}
where $N$ represents the number of pixels. This weighting reduces the influence of samples with high entropy, which indicates low confidence and potential to mislead the network. The uncertainty-based pixel-level weighting mechanisms are
\begin{equation}
    U(PL_T)= (2PL_T-1)^2, \ \ \ U(PL_F)= (2PL_F-1)^2,
\end{equation}
which attenuates pixels with predictions near 0.5, reflecting ambiguity in distinguishing foreground from background. 

Unlike prior methods~\cite{he2023weaklysupervised,he2025segment} that simply combine these, we define a basic weighted loss ${R}_b(\bigcdot)$ and apply it piecewise based on sample difficulty. The basic loss ${R}_b(\bigcdot)$ combines sample confidence $(1-E(\bigcdot))$ with pixel certainty $U(\bigcdot)$:
\begin{equation}\hspace{-3mm}
\begin{aligned}
        R_b(PL_T,\hat{{Y}}_1)=& (1-E(PL_T))\cdot U(PL_T) \cdot \\ 
        &(L_{ce}(\hat{{Y}}_1,PL_T)+L_{IoU}(\hat{{Y}}_1,PL_T)),
\end{aligned}
\end{equation}
where $L_{ce}(\bigcdot)$ and $L_{IoU}(\bigcdot)$ correspond to the cross-entropy loss and the intersection-over-union loss, respectively. $\hat{{Y}}_1$ is the output of the student segmenter, defined as
\begin{equation}
\hat{{Y}}_1=\text{Seg}_S(Aug_w({X})).
\end{equation}
We further address two extreme cases: \\
\textbf{(1) Difficult samples}: If the entropy $E(PL_T)$ is greater than or equal to 0.8, indicating low confidence, we introduce a mask $w_{PL_T}$ to isolate trustworthy pixels:
\begin{equation}
\begin{aligned}
    w_{PL_T}=&
\left\{
\begin{array}{ll}
    0  &PL_T \in [0.1,0.9],\\
    1  & \text{Otherwise}.
\end{array}\right.
\end{aligned}
\end{equation}
In this case, the refined pseudo-labels are defined as:
\begin{equation}
    R(PL_T,\hat{{Y}}_1)= w_{PL_T}\cdot R_b(PL_T,\hat{{Y}}_1). 
\end{equation}
\begin{table*}[t]
\begin{minipage}{\textwidth}
	\centering
	\setlength{\abovecaptionskip}{0cm}
	\caption{Results on COD of the WSCOS task with point supervision and scribble supervision. 
		The best two results are in {\color[HTML]{FF0000} \textbf{red}} and {\color[HTML]{00B0F0} \textbf{blue}} fonts.}
	\resizebox{\textwidth}{!}{
		\setlength{\tabcolsep}{1mm}
		\begin{tabular}{l|c|cccc|cccc|cccc|cccc}
			\toprule
			\multicolumn{1}{l|}{} & \multicolumn{1}{c|}{} & \multicolumn{4}{c|}{\textit{CHAMELEON} } & \multicolumn{4}{c|}{\textit{CAMO}} & \multicolumn{4}{c|}{\textit{COD10K}} & \multicolumn{4}{c}{\textit{NC4K}} \\ \cline{3-18}
			\multicolumn{1}{l|}{\multirow{-2}{*}{Methods}} & \multicolumn{1}{c|}{\multirow{-2}{*}{Pub.}} & {\cellcolor{gray!40}$M$~$\downarrow$} &{\cellcolor{gray!40}$F_\beta$~$\uparrow$} &{\cellcolor{gray!40}$E_\phi$~$\uparrow$} & \multicolumn{1}{c|}{\cellcolor{gray!40}$S_\alpha$~$\uparrow$}& {\cellcolor{gray!40}$M$~$\downarrow$} &{\cellcolor{gray!40}$F_\beta$~$\uparrow$} &{\cellcolor{gray!40}$E_\phi$~$\uparrow$} & \multicolumn{1}{c|}{\cellcolor{gray!40}$S_\alpha$~$\uparrow$}& {\cellcolor{gray!40}$M$~$\downarrow$} &{\cellcolor{gray!40}$F_\beta$~$\uparrow$} &{\cellcolor{gray!40}$E_\phi$~$\uparrow$} & \multicolumn{1}{c|}{\cellcolor{gray!40}$S_\alpha$~$\uparrow$}& {\cellcolor{gray!40}$M$~$\downarrow$} &{\cellcolor{gray!40}$F_\beta$~$\uparrow$} &{\cellcolor{gray!40}$E_\phi$~$\uparrow$} & \multicolumn{1}{c}{\cellcolor{gray!40}$S_\alpha$~$\uparrow$}\\ \midrule
			\multicolumn{18}{c}{Scribble Supervision} \\ \midrule
			SAM~\cite{kirillov2023segment}                                         & \multicolumn{1}{c|}{ICCV23}                              & 0.207                                 & 0.595                                 & 0.647                                 & 0.635                                 & 0.160                                 & 0.597                                 & 0.639                                 & 0.643                                 & 0.093                                 & 0.673                                 & 0.737                                 & 0.730                                 & 0.118                                 & 0.675                                 & 0.723                                 & 0.717                                 \\
		 $\text{SAM-W}_\text{S}$~\cite{chen2023sam}                                         & \multicolumn{1}{c|}{ICCVW23}                              & 0.069                & 0.751                & 0.835                & 0.661                & 0.097                & 0.696                & 0.788                & 0.738                & 0.049                & 0.712                & 0.833                & 0.770                & 0.066                & 0.757                & 0.842                & 0.768                \\
			WSSA~\cite{zhang2020weakly}                                          & CVPR20                                             & 0.067                                 & 0.692                                 & 0.860                                 & 0.782                                 & 0.118                                 & 0.615                                 & 0.786                                 & 0.696                                 & 0.071                                 & 0.536                                 & 0.770                                 & 0.684                                 & 0.091                                 & 0.657                                 & 0.779                                 & 0.761                                 \\
			SCWS~\cite{yu2021structure}                                       & AAAI21                                             & 0.053                                 & 0.758                                 & 0.881                                 & 0.792                                 & 0.102                                 & 0.658                                 & 0.795                                 & 0.713                                 & 0.055                                 & 0.602                                 & 0.805                                 & 0.710                                 & 0.073                                 & 0.723                                 & 0.814                                 & 0.784                                 \\
			TEL~\cite{liang2022tree}                                           & CVPR22                                             & 0.073                                 & 0.708                                 & 0.827                                 & 0.785                                 & 0.104                                 & 0.681                                 & 0.797                                 & 0.717                                 & 0.057                                 & 0.633                                 & 0.826                                 & 0.724                                 & 0.075                                 & 0.754                                 & 0.832                                 & 0.782                                 \\
			SCOD~\cite{he2022weakly}                                         & AAAI23                                             & 0.046                                 & {\color[HTML]{FF0000} \textbf{0.791}}                                 & 0.897                                 & 0.818                                 & 0.092                                 & 0.709                                 & 0.815                                 & 0.735                                 & 0.049                                 & 0.637                                 & 0.832                                 & 0.733                                 & 0.064                                 & 0.751                                 & 0.853                                 & 0.779                                 \\
      GenSAM~\cite{hu2024relax}   & AAAI24 & 0.090 & 0.680 & 0.807 & 0.764 & 0.113 & 0.659 & 0.775 & 0.719 & 0.067 & 0.681 & 0.838 & 0.775 & 0.097 & 0.687 & 0.750 & 0.732 \\
WS-SAM~\cite{he2023weaklysupervised} & NIPS23 & 0.046                                 & 0.777                                 & 0.897                                 & {0.824} & 0.092                                 & {0.742} & 0.818                                 & {0.759} & {0.038} & {0.719} & {0.878} & {0.803} & {0.052} & {0.802} & {0.886} & {0.829} \\ 
   SEE~\cite{he2025segment} & \multicolumn{1}{c|}{TPAMI25}& {\color[HTML]{00B0F0} \textbf{0.044}} & 0.785                                 & {\color[HTML]{00B0F0} \textbf{0.903}} & {\color[HTML]{00B0F0} \textbf{0.826}} & {\color[HTML]{00B0F0} \textbf{0.090}} & {\color[HTML]{00B0F0} \textbf{0.747}} & {\color[HTML]{00B0F0} \textbf{0.826}} & {\color[HTML]{00B0F0} \textbf{0.765}} & {\color[HTML]{00B0F0} \textbf{0.036}} & {\color[HTML]{00B0F0} \textbf{0.729}} & {\color[HTML]{FF0000} \textbf{0.883}} & {\color[HTML]{00B0F0} \textbf{0.807}} & {\color[HTML]{00B0F0} \textbf{0.051}} & {\color[HTML]{00B0F0} \textbf{0.808}} & {\color[HTML]{00B0F0} \textbf{0.891}} & {\color[HTML]{FF0000} \textbf{0.836}} \\
   \rowcolor{c2!20} SCALER (Ours) & \multicolumn{1}{c|}{---} &{\color[HTML]{FF0000} \textbf{0.042}} & {\color[HTML]{00B0F0} \textbf{0.788}} & {\color[HTML]{FF0000} \textbf{0.912}} & {\color[HTML]{FF0000} \textbf{0.838}} & {\color[HTML]{FF0000} \textbf{0.089}} & {\color[HTML]{FF0000} \textbf{0.785}} & {\color[HTML]{FF0000} \textbf{0.834}} & {\color[HTML]{FF0000} \textbf{0.778}} & {\color[HTML]{FF0000} \textbf{0.034}} & {\color[HTML]{FF0000} \textbf{0.736}} & {\color[HTML]{00B0F0} \textbf{0.882}} & {\color[HTML]{FF0000} \textbf{0.815}} & {\color[HTML]{FF0000} \textbf{0.050}} & {\color[HTML]{FF0000} \textbf{0.809}} & {\color[HTML]{FF0000} \textbf{0.897}} & {\color[HTML]{00B0F0} \textbf{0.817}}   \\ 		
   \midrule
			\multicolumn{18}{c}{Point Supervision} \\ \midrule
			SAM~\cite{kirillov2023segment} & \multicolumn{1}{c|}{ICCV23} & 0.207 & 0.595                                 & 0.647                                 & 0.635                                 & 0.160                                 & 0.597                                 & 0.639                                 & 0.643                                 & 0.093                                 & 0.673                                 & 0.737                                 & 0.730                                 & 0.118                                 & 0.675                                 & 0.723                                 & 0.717                                 \\
		$\text{SAM-W}_\text{P}$~\cite{chen2023sam} & \multicolumn{1}{c|}{ICCVW23} & 0.095 & 0.708 & 0.752 & 0.688 & 0.117 & 0.653 & 0.698 & 0.681 & 0.068 & 0.697 & 0.805 & 0.767 & 0.078 & 0.736 & 0.793 & 0.782 \\
			WSSA~\cite{zhang2020weakly}                                          & CVPR20                                             & 0.105                                 & 0.660                                 & 0.712                                 & 0.711                                 & 0.148                                 & 0.607                                 & 0.652                                 & 0.649                                 & 0.087                                 & 0.509                                 & 0.733                                 & 0.642                                 & 0.104                                 & 0.688                                 & 0.756                                 & 0.743                                 \\
			SCWS~\cite{yu2021structure}                                       & AAAI21                                             & 0.097                                 & 0.684                                 & 0.739                                 & 0.714                                 & 0.142                                 & 0.624                                 & 0.672                                 & {{0.687}}                                 & 0.082                                 & 0.593                                 & 0.777                                 & 0.738                                 & 0.098                                 & 0.695                                 & 0.767                                 & 0.754                                 \\
			TEL~\cite{liang2022tree}                                           & CVPR22                                             & 0.094                                 & {{0.712}}                                 & 0.751                                 & {{0.746}}                                 & 0.133                                 & {{0.662}}                                 & 0.674                                 & 0.645                                 & 0.063                                 & 0.623                                 & 0.803                                 & 0.727                                 & 0.085                                 & 0.725                                 & 0.795                                 & 0.766                                 \\
			SCOD~\cite{he2022weakly}                                         & AAAI23                                             & 0.092 & 0.688                                 & 0.746                                 & 0.725                                 & 0.137                                 & 0.629                                 & 0.688                                 & 0.663                                 & 0.060                                 & 0.607                                 & 0.802                                 & 0.711                                 & 0.080                                 & 0.744                                 & 0.796                                 & 0.758                                 \\
   GenSAM~\cite{hu2024relax}   &   AAAI24    & 0.090 & 0.680 & 0.807 & 0.764 & 0.113 & 0.659 & 0.775 & 0.719 & 0.067 & 0.681 & 0.838 & 0.775 & 0.097 & 0.687 & 0.750 & 0.732 \\
		WS-SAM~\cite{he2023weaklysupervised}  & NIPS23 & {0.056} & {0.767} & {0.868} & {0.805} & {0.102} &{0.703} & {0.757} & {0.718} & {0.039}& {0.698} & {0.856} & {0.790} & {0.057} & {0.801} & {0.859} & {0.813} \\ 
   SEE~\cite{he2025segment} & \multicolumn{1}{c|}{TPAMI25}& {\color[HTML]{00B0F0} \textbf{0.055}} & {\color[HTML]{00B0F0} \textbf{0.772}} & {\color[HTML]{FF0000} \textbf{0.872}} & {\color[HTML]{00B0F0} \textbf{0.806}} & {\color[HTML]{00B0F0} \textbf{0.098}} & {\color[HTML]{00B0F0} \textbf{0.712}} & {\color[HTML]{00B0F0} \textbf{0.769}} & {\color[HTML]{00B0F0} \textbf{0.721}} & {\color[HTML]{FF0000} \textbf{0.038}} & {\color[HTML]{00B0F0} \textbf{0.706}} & {\color[HTML]{00B0F0} \textbf{0.862}} & {\color[HTML]{00B0F0} \textbf{0.796}} & {\color[HTML]{00B0F0} \textbf{0.055}} & {\color[HTML]{00B0F0} \textbf{0.806}} & {\color[HTML]{FF0000} \textbf{0.867}} & {\color[HTML]{00B0F0} \textbf{0.817}}\\
\rowcolor{c2!20} SCALER (Ours) & \multicolumn{1}{c|}{---} & {\color[HTML]{FF0000} \textbf{0.054}} & {\color[HTML]{FF0000} \textbf{0.779}} & {\color[HTML]{00B0F0} \textbf{0.869}} & {\color[HTML]{FF0000} \textbf{0.809}} & {\color[HTML]{FF0000} \textbf{0.096}} & {\color[HTML]{FF0000} \textbf{0.716}} & {\color[HTML]{FF0000} \textbf{0.779}} & {\color[HTML]{FF0000} \textbf{0.729}} & {\color[HTML]{00B0F0} \textbf{0.039}} & {\color[HTML]{FF0000} \textbf{0.718}} & {\color[HTML]{FF0000} \textbf{0.879}} & {\color[HTML]{FF0000} \textbf{0.802}} & {\color[HTML]{FF0000} \textbf{0.054}} & {\color[HTML]{FF0000} \textbf{0.809}} & {\color[HTML]{00B0F0} \textbf{0.863}} & {\color[HTML]{FF0000} \textbf{0.821}}  \\ 		\bottomrule
	\end{tabular}}
	\label{table:CODWeak}
	\vspace{-0.4cm}
\end{minipage} 
\end{table*}
\textbf{(2) Simple samples}: For samples where the entropy of predictions $E(\hat{{Y}}_1)$ is less than or equal to 0.2, we apply additional strong augmentations $Aug_s(\cdot)$ to increase task difficulty, obtaining another prediction $\hat{{Y}}'_1 = \text{Seg}_S(Aug_s({X}))$. The refinement strategy becomes:
\begin{equation}
R(PL_T,\hat{{Y}}_1)=R_b(PL_T,\hat{{Y}}_1)+R_b(PL_T,\hat{{Y}}'_1).
\end{equation}
Overall, the piecewise refinement strategy is defined as:
\begin{equation}\hspace{-3mm}
\begin{array}{c}
R(PL_T, \hat{{Y}}_1)= \\
\begin{cases}
  w_{PL_F}\cdot R_b(PL_T,\hat{{Y}}_1), & E(PL_T)\!\ge\!0.8,\\
  R_b(PL_T,\hat{{Y}}_1)\!+\!R_b(PL_T,\hat{{Y}}'_1), & E(\hat{{Y}}_1)\!\le\!0.2,\\[3pt]
  R_b(PL_T,\hat{{Y}}_1), & \text{Otherwise.}
\end{cases}
\end{array}
\end{equation}

\noindent \textbf{Optimization}.
Under weak supervision, the segmenter is trained with the annotations ${Y}$ and our refinement strategy:
\begin{equation}
    \mathcal{L}_w^1=R(PL_T,\hat{{Y}}_1)+R(PL_F,\hat{{Y}}_1)+L_{pce}({Y},\hat{{Y}}_1),
\end{equation}
where $\mathcal{L}_{pce}$ is the partial cross-entropy loss. Under semi-supervision, $\mathcal{L}_{pce}$ is replaced by $\mathcal{L}_{ce}+\mathcal{L}_{IoU}$ for labeled samples, while unlabeled samples use the refinement loss:
\begin{equation}
\begin{aligned}
        \mathcal{L}_s^1=&(R(PL_T,\hat{{Y}}_1)+R(PL_F,\hat{{Y}}_1)) \\
        &+(L_{ce}({Y},\hat{{Y}}_1)+L_{IoU}({Y},\hat{{Y}}_1)).
\end{aligned}
\end{equation}
The teacher weights $\theta_T$ are updated via an exponential moving average (EMA) strategy from the student $\theta_S$:
\begin{equation}\label{eq:EMA}
\theta_t=\eta \theta_T + (1-\eta)\theta_S, \ \ \text{with } \ \eta=0.996.
\end{equation}

\subsection{Phase \uppercase\expandafter{\romannumeral2}: Optimizing SAM via Seg. Feedback }
This phase enables the novel \textit{bi-directional} learning. We reverse the roles: the specialized mean-teacher framework ($\text{Seg}_S, \text{Seg}_T$) is now \textit{fixed} and serves as a ``specialist'' teacher to update the \textit{learnable} SAM (via SAMadapter~\cite{chen2023sam}).

This requires SAM (a generalist) to learn from a specialist's noisy pseudo-labels. We generate a consensus pseudo-label ${PL}_M$ from the fixed segmenter framework, ${PL}_M = ({PL}_S + {PL}_{T}) / 2$, where ${PL}_S = \text{Seg}_S(Aug_w({X}))$ and ${PL}_T = \text{Seg}_T(Aug_w({X}))$. We then design two losses to make this knowledge transfer robust.

\noindent\textbf{Augmentation Invariance Loss ($\mathcal{L}_{ai}$)}. We exploit SAM's robustness by enforcing prediction consistency between weak and strong augmentations:
\begin{equation}
    \mathcal{L}_{ai}=\mathcal{L}_{ce}(\hat{{Y}}_2,\hat{{Y}}'_2)+\mathcal{L}_{IoU}(\hat{{Y}}_2,\hat{{Y}}'_2),
\end{equation}
where $\hat{{Y}}_2=\text{SAM}(Aug_w({X}))$ and $\hat{{Y}}'_2=\text{SAM}(Aug_s({X}))$. This loss function facilitates robust representation learning while not impairing generalization.

\noindent\textbf{Noise Resistance Loss ($\mathcal{L}_{nr}$)}. This loss distills knowledge from the segmenter's pseudo-label ${PL}_M$ while protecting SAM from its noise. We apply the pixel-level uncertainty weight $U(\cdot)$ to ignore ambiguous pixels, but we omit the sample-level entropy weight $E(\cdot)$ to ensure SAM learns from all samples, even difficult ones. $\mathcal{L}_{nr}$ is defined as:
\begin{equation}\hspace{-3mm}
    \mathcal{L}_{nr}\!=\! U(PL_M)\cdot(\mathcal{L}_{ce}(\hat{{Y}}_2,PL_M)\!+\!\mathcal{L}_{IoU}(\hat{{Y}}_2,PL_M)),
\end{equation}
The two loss functions, by increasing the diversity and quantity of training samples, jointly mitigate the overfitting that SAM may encounter under limited supervision. 

\noindent \textbf{Optimization}.
SAM, with weak supervision, is updated via
\begin{equation}
\mathcal{L}_w^2=\mathcal{L}_{ai}+\mathcal{L}_{nr}+\mathcal{L}_{pce}({Y},\hat{{Y}}_2).
\end{equation}
Under semi-supervision, the loss $\mathcal{L}_s^2$ combines this with a standard supervised loss on labeled data, formulated as
\begin{equation}
    \mathcal{L}_s^2=[\mathcal{L}_{ai}+\mathcal{L}_{nr}]+[\mathcal{L}_{ce}({Y},\hat{{Y}}_2)+\mathcal{L}_{IoU}({Y},\hat{{Y}}_2)].
\end{equation}
\begin{table*}[t]
\begin{minipage}{\textwidth}
	\centering
	\setlength{\abovecaptionskip}{0cm}
	\caption{Results on COD of the SSCOS task with 1/8 and 1/16 labeled training data. 
  }
	\resizebox{\textwidth}{!}{
		\setlength{\tabcolsep}{1mm}
		\begin{tabular}{l|c|cccc|cccc|cccc|cccc}
			\toprule
			\multicolumn{1}{l|}{} & \multicolumn{1}{c|}{} & \multicolumn{4}{c|}{\textit{CHAMELEON} } & \multicolumn{4}{c|}{\textit{CAMO}} & \multicolumn{4}{c|}{\textit{COD10K}} & \multicolumn{4}{c}{\textit{NC4K}} \\ \cline{3-18}
			\multicolumn{1}{l|}{\multirow{-2}{*}{Methods}} & \multicolumn{1}{c|}{\multirow{-2}{*}{Pub.}} & {\cellcolor{gray!40}$M$~$\downarrow$} &{\cellcolor{gray!40}$F_\beta$~$\uparrow$} &{\cellcolor{gray!40}$E_\phi$~$\uparrow$} & \multicolumn{1}{c|}{\cellcolor{gray!40}$S_\alpha$~$\uparrow$}& {\cellcolor{gray!40}$M$~$\downarrow$} &{\cellcolor{gray!40}$F_\beta$~$\uparrow$} &{\cellcolor{gray!40}$E_\phi$~$\uparrow$} & \multicolumn{1}{c|}{\cellcolor{gray!40}$S_\alpha$~$\uparrow$}& {\cellcolor{gray!40}$M$~$\downarrow$} &{\cellcolor{gray!40}$F_\beta$~$\uparrow$} &{\cellcolor{gray!40}$E_\phi$~$\uparrow$} & \multicolumn{1}{c|}{\cellcolor{gray!40}$S_\alpha$~$\uparrow$}& {\cellcolor{gray!40}$M$~$\downarrow$} &{\cellcolor{gray!40}$F_\beta$~$\uparrow$} &{\cellcolor{gray!40}$E_\phi$~$\uparrow$} & \multicolumn{1}{c}{\cellcolor{gray!40}$S_\alpha$~$\uparrow$}\\ \midrule
			\multicolumn{18}{c}{1/8 Labeled Training Data} \\ \midrule
			SAM~\cite{kirillov2023segment}        & ICCV23       & 0.207                                 & 0.595                                 & 0.647                                 & 0.635                                 & 0.160                                 & 0.597                                 & 0.639                                 & 0.643                                 & 0.093                                 & 0.673                                 & 0.737                                 & 0.730                                 & 0.118                                 & 0.675                                 & 0.723                                 & 0.717                                 \\
SAM-S~\cite{chen2023sam}      &  ICCVW23   & 0.136                                 & 0.667                                 & 0.695                                 & 0.672                                 & 0.133                                 & 0.637                                 & 0.678                                 & 0.662                                 & 0.073                                 & {0.695} & 0.770                                 & 0.751                                 & 0.085                                 & 0.715                                 & 0.756                                 & 0.767                                 \\
DTEN~\cite{ren2023towards}       & CVPR23 & 0.062                                 & 0.775                                 & 0.862                                 & 0.803                                 & 0.103                                 & 0.688                                 & 0.781                                 & 0.742                                 & 0.054                                 & 0.635                                 & 0.791                                 & 0.747                                 & 0.070                                 & 0.733                                 & 0.833                                 & 0.790                                 \\
PGCL~\cite{basak2023pseudo}       & CVPR23  & 0.051                                 & 0.792                                 & 0.878                                 & 0.833                                 & 0.096                                 & 0.705                                 & 0.803                                 & 0.755                                 & 0.051                                 & 0.658                                 & 0.798                                 & 0.752                                 & 0.063                                 & 0.753                                 & 0.838                                 & 0.803                                 \\
EPS~\cite{lee2023saliency}        & TPAMI23 & 0.042                                 & 0.810                                 & 0.891                                 & 0.862                                 & 0.090                                 & 0.721                                 & 0.815                                 & 0.753                                 & 0.047                                 & 0.659                                 & 0.806                                 & 0.761                                 & 0.058                                 & 0.765                                 & 0.855                                 & 0.809                                 \\
CoSOD~\cite{chakraborty2024unsupervised}      & WACV24  & 0.047                                 & 0.802                                 & 0.883                                 & 0.850                                 & 0.092                                 & 0.730                                 & 0.822                                 & 0.761                                 & {{0.046}}                                 & 0.673                                 & 0.813                                 & 0.767                                 & 0.061                                 & 0.764                                 & 0.862                                 & 0.818                                 \\
SEE~\cite{he2025segment} &   TPAMI25   & {\color[HTML]{00B0F0} \textbf{0.035}} & {\color[HTML]{00B0F0} \textbf{0.825}} & {\color[HTML]{00B0F0} \textbf{0.903}} & {\color[HTML]{00B0F0} \textbf{0.873}} & {\color[HTML]{00B0F0} \textbf{0.083}} & {\color[HTML]{00B0F0} \textbf{0.753}} & {\color[HTML]{FF0000} \textbf{0.843}} & {\color[HTML]{00B0F0} \textbf{0.776}} & {\color[HTML]{FF0000} \textbf{0.040}} & {\color[HTML]{00B0F0} \textbf{0.703}} & {\color[HTML]{00B0F0} \textbf{0.839}} & {\color[HTML]{FF0000} \textbf{0.786}} & {\color[HTML]{00B0F0} \textbf{0.053}} & {\color[HTML]{00B0F0} \textbf{0.778}} & {\color[HTML]{FF0000} \textbf{0.889}} & {\color[HTML]{00B0F0} \textbf{0.839}} \\
\rowcolor{c2!20} SCALER (Ours) &   --- &{\color[HTML]{FF0000} \textbf{0.033}} & {\color[HTML]{FF0000} \textbf{0.837}} & {\color[HTML]{FF0000} \textbf{0.911}} & {\color[HTML]{FF0000} \textbf{0.878}} & {\color[HTML]{FF0000} \textbf{0.081}} & {\color[HTML]{FF0000} \textbf{0.757}} & {\color[HTML]{00B0F0} \textbf{0.835}} & {\color[HTML]{FF0000} \textbf{0.779}} & {\color[HTML]{FF0000} \textbf{0.040}} & {\color[HTML]{FF0000} \textbf{0.714}} & {\color[HTML]{FF0000} \textbf{0.845}} & {\color[HTML]{00B0F0} \textbf{0.784}} & {\color[HTML]{FF0000} \textbf{0.052}} & {\color[HTML]{FF0000} \textbf{0.783}} & {\color[HTML]{00B0F0} \textbf{0.885}} & {\color[HTML]{FF0000} \textbf{0.851}} \\ 
   \midrule
			\multicolumn{18}{c}{1/16 Labeled Training Data} \\ \midrule
			SAM~\cite{kirillov2023segment}        & ICCV23 & 0.207                                 & 0.595                                 & 0.647                                 & 0.635                                 & 0.160                                 & 0.597                                 & 0.639                                 & 0.643                                 & 0.093                                 & 0.673                                 & 0.737                                 & 0.730                                 & 0.118                                 & 0.675                                 & 0.723                                 & 0.717                                 \\
SAM-S~\cite{chen2023sam}      &    ICCVW23   & 0.150                                 & 0.642                                 & 0.680                                 & 0.657                                 & 0.146                                 & 0.624                                 & 0.663                                 & 0.647                                 & 0.078                                 & {\color[HTML]{00B0F0} \textbf{0.682}} & 0.752                                 & 0.738                                 & 0.093                                 & 0.692                                 & 0.741                                 & 0.753                                 \\
DTEN~\cite{ren2023towards}       & CVPR23               & 0.070                                 & 0.731                                 & 0.827                                 & 0.776                                 & 0.123                                 & 0.665                                 & 0.765                                 & 0.718                                 & 0.080                                 & 0.618                                 & 0.762                                 & 0.714                                 & 0.083                                 & 0.704                                 & 0.797                                 & 0.772                                 \\
PGCL~\cite{basak2023pseudo}       & CVPR23                & 0.057                                 & 0.752                                 & 0.850                                 & 0.801                                 & 0.116                                 & 0.682                                 & 0.782                                 & 0.722                                 & 0.061                                 & 0.637                                 & 0.779                                 & 0.728                                 & 0.071                                 & 0.719                                 & 0.809                                 & 0.789                                 \\
EPS~\cite{lee2023saliency}        & TPAMI23               & 0.049                                 & 0.763                                 & 0.843                                 & 0.828                                 & 0.103                                 & 0.697                                 & 0.796                                 & 0.735                                 & 0.056                                 & 0.646                                 & 0.787                                 & 0.736                                 & 0.066                                 & 0.737                                 & 0.833                                 & 0.801                                 \\
CoSOD~\cite{chakraborty2024unsupervised}      & WACV24                & 0.055                                 & 0.758                                 & 0.856                                 & 0.830                                 & 0.099                                 & 0.702                                 & 0.793                                 & 0.730                                 & 0.055                                 & 0.650                                 & 0.795                                 & 0.740                                 & 0.070                                 & 0.726                                 & 0.825                                 & 0.792                                 \\
SEE~\cite{he2025segment} &   TPAMI25     & {\color[HTML]{00B0F0} \textbf{0.040}} & {\color[HTML]{00B0F0} \textbf{0.793}} & {\color[HTML]{FF0000} \textbf{0.885}} & {\color[HTML]{00B0F0} \textbf{0.852}} & {\color[HTML]{00B0F0} \textbf{0.093}} & {\color[HTML]{00B0F0} \textbf{0.716}} & {\color[HTML]{00B0F0} \textbf{0.810}} & {\color[HTML]{00B0F0} \textbf{0.747}} & {\color[HTML]{00B0F0} \textbf{0.046}} & {0.679} & {\color[HTML]{00B0F0} \textbf{0.803}} & {\color[HTML]{00B0F0} \textbf{0.745}} & {\color[HTML]{00B0F0} \textbf{0.060}} & {\color[HTML]{00B0F0} \textbf{0.757}} & {\color[HTML]{FF0000} \textbf{0.863}} & {\color[HTML]{00B0F0} \textbf{0.812}} \\ 
\rowcolor{c2!20} SCALER (Ours) &   --- &{\color[HTML]{FF0000} \textbf{0.039}} & {\color[HTML]{FF0000} \textbf{0.799}} & {\color[HTML]{00B0F0} \textbf{0.878}} & {\color[HTML]{FF0000} \textbf{0.856}} & {\color[HTML]{FF0000} \textbf{0.091}} & {\color[HTML]{FF0000} \textbf{0.725}} & {\color[HTML]{FF0000} \textbf{0.831}} & {\color[HTML]{FF0000} \textbf{0.756}} & {\color[HTML]{FF0000} \textbf{0.044}} & {\color[HTML]{FF0000} \textbf{0.685}} & {\color[HTML]{FF0000} \textbf{0.821}} & {\color[HTML]{FF0000} \textbf{0.764}} & {\color[HTML]{FF0000} \textbf{0.058}} & {\color[HTML]{FF0000} \textbf{0.764}} & {\color[HTML]{00B0F0} \textbf{0.858}} & {\color[HTML]{FF0000} \textbf{0.824}} \\ \bottomrule
	\end{tabular}}
	\label{table:CODSemi}
	\vspace{-0.4cm}
\end{minipage}
\end{table*}
\begin{table*}[t]
\begin{minipage}{\textwidth}
\centering
\setlength{\abovecaptionskip}{0cm} 	\setlength{\belowcaptionskip}{0cm}
\includegraphics[width=\linewidth]{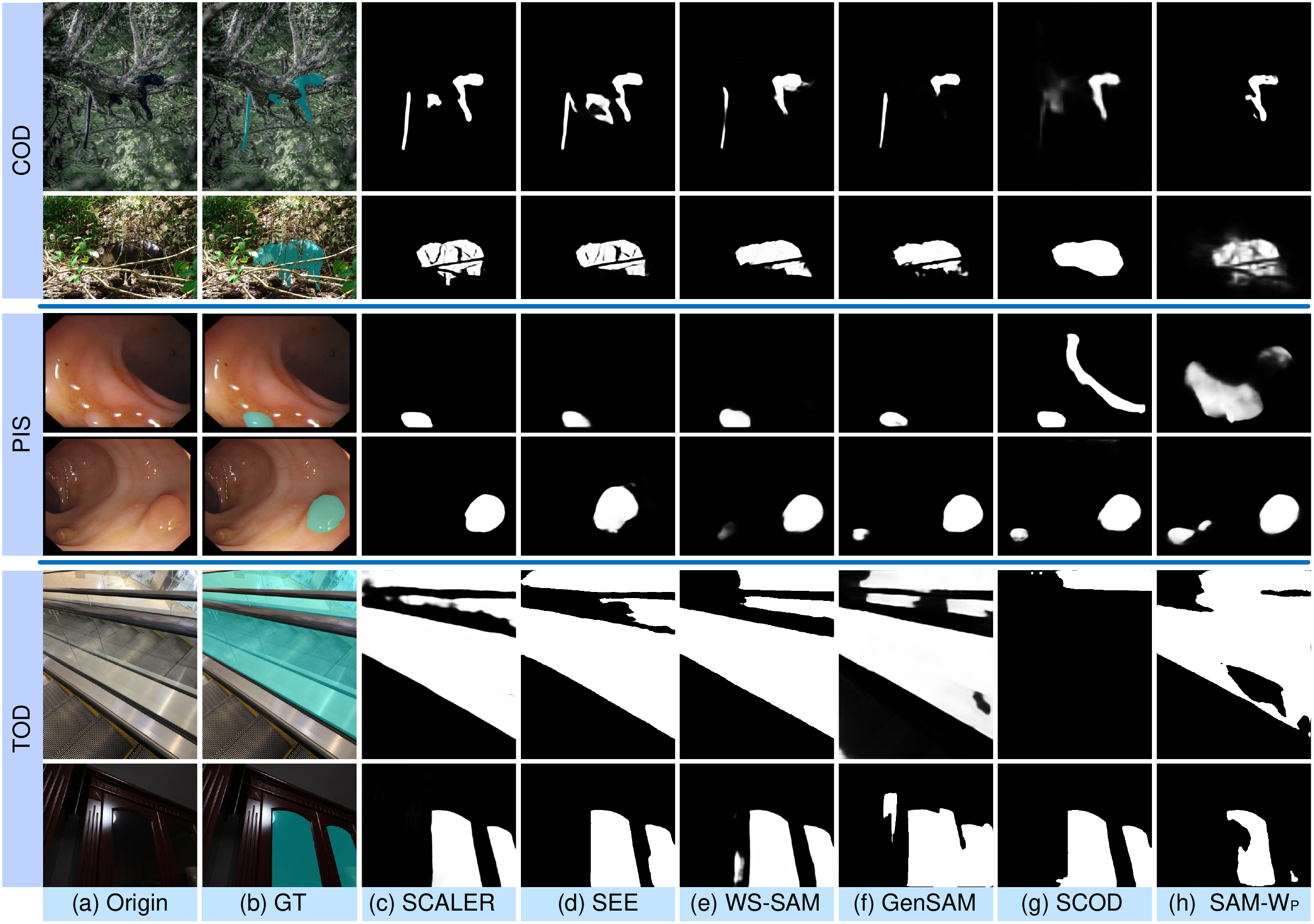}  
	\captionof{figure}{Visualizations for COS tasks with point supervision.}
    \label{fig:COSQuali}
	\vspace{-6mm}
\end{minipage}

\end{table*}

\begin{table*}[t]
 \begin{minipage}{\textwidth}
\centering
	\setlength{\abovecaptionskip}{0cm}
	\caption{Results on PIS and TOD of the WSCOS task with point supervision. 
	}
	\resizebox{\textwidth}{!}{
		\setlength{\tabcolsep}{0.8mm}
		\begin{tabular}{l|cccc|cccc|cccc|cccc|cccc}
			\toprule
			& \multicolumn{12}{c|}{Polyp Image Segmentation (PIS)}& \multicolumn{8}{c}{Transparent Object Detection (TOD)}\\ \cline{2-21}
			& \multicolumn{4}{c|}{\textit{CVC-ColonDB}}& \multicolumn{4}{c|}{\textit{ETIS}}& \multicolumn{4}{c|}{\textit{Kvasir}}& \multicolumn{4}{c|}{\textit{GDD}}& \multicolumn{4}{c}{\textit{GSD}}\\ \cline{2-21}
			\multirow{-3}{*}{Methods} & {\cellcolor{gray!40}$M$~$\downarrow$} &{\cellcolor{gray!40}$F_\beta$~$\uparrow$} &{\cellcolor{gray!40}$E_\phi$~$\uparrow$} & \multicolumn{1}{c|}{\cellcolor{gray!40}$S_\alpha$~$\uparrow$}& {\cellcolor{gray!40}$M$~$\downarrow$} &{\cellcolor{gray!40}$F_\beta$~$\uparrow$} &{\cellcolor{gray!40}$E_\phi$~$\uparrow$} & \multicolumn{1}{c|}{\cellcolor{gray!40}$S_\alpha$~$\uparrow$}& {\cellcolor{gray!40}$M$~$\downarrow$} &{\cellcolor{gray!40}$F_\beta$~$\uparrow$} &{\cellcolor{gray!40}$E_\phi$~$\uparrow$} & \multicolumn{1}{c|}{\cellcolor{gray!40}$S_\alpha$~$\uparrow$}& {\cellcolor{gray!40}$M$~$\downarrow$} &{\cellcolor{gray!40}$F_\beta$~$\uparrow$} &{\cellcolor{gray!40}$E_\phi$~$\uparrow$} & \multicolumn{1}{c|}{\cellcolor{gray!40}$S_\alpha$~$\uparrow$}& {\cellcolor{gray!40}$M$~$\downarrow$} &{\cellcolor{gray!40}$F_\beta$~$\uparrow$} &{\cellcolor{gray!40}$E_\phi$~$\uparrow$} & \multicolumn{1}{c}{\cellcolor{gray!40}$S_\alpha$~$\uparrow$}\\ \midrule
			SAM~\cite{kirillov2023segment}& 0.479                                 & 0.343                                 & 0.419                                 & 0.427                                 & 0.429                                 & 0.439                                 & 0.512                                 & 0.503                                 & 0.320                                 & 0.545                                 & 0.564                                 & 0.582                                 & 0.245                                 & 0.512                                 & 0.530                                 & 0.551                                 & 0.266                                 & 0.473                                 & 0.501                                 & 0.514                                 \\
			$\text{SAM-W}_\text{P}$~\cite{chen2023sam}& 0.146 & 0.612 & 0.690 & 0.683 & 0.125 & 0.650 & 0.731 & 0.728 & 0.093 & 0.818 & 0.823 & 0.815 & 0.143 & 0.691 & 0.733 & 0.636 & 0.137 & 0.715 & 0.752 & 0.681 \\
			WSSA~\cite{zhang2020weakly} & 0.127                                 & 0.645                                 & 0.732                                 & 0.713                                 & 0.123                                 & 0.647                                 & 0.733                                 & 0.762                                 & 0.082                                 & 0.822                                 & 0.852                                 & 0.828                                 & 0.173                                 & 0.652                                 & 0.710                                 & 0.616                                 & 0.185                                 & 0.661                                 & 0.712                                 & 0.650                                 \\
			SCWS~\cite{yu2021structure}                      & 0.082                                 & 0.674                                 & 0.758                                 & 0.787                                 & 0.085                                 & 0.646                                 & 0.768                                 & 0.731                                 & 0.078                                 & 0.837                                 & 0.860                                 & 0.831                                 & 0.170                                 & 0.631                                 & 0.702                                 & 0.613                                 & 0.172                                 & 0.706                                 & 0.738                                 & 0.673                                 \\
			TEL~\cite{liang2022tree}                       & 0.089                                 & 0.669                                 & 0.743                                 & 0.761                                 & 0.083                                 & 0.639                                 & 0.776                                 & 0.726                                 & 0.091                                 & 0.810                                 & 0.826                                 & 0.804                                 & 0.230                                 & 0.640                                 & 0.586                                 & 0.536                                 & 0.275                                 & 0.571                                 & 0.501                                 & 0.495                                 \\
			SCOD~\cite{he2022weakly}                      & 0.077                                 & 0.691                                 & 0.795                                 & 0.802                                 & 0.071                                 & 0.664                                 & 0.802                                 & 0.766                                 & 0.071                                 & 0.853                                 & 0.877                                 & {{0.836}} & 0.146                                 & 0.801                                 & 0.778                                 & 0.723                                 & 0.154                                 & 0.743                                 & 0.751                                 & 0.710                                 \\
   GenSAM~\cite{hu2024relax}& 0.079 & 0.665 & 0.739 & 0.793 & 0.073 & 0.665 & 0.813 & 0.772 & 0.073 & 0.841 & 0.866 & 0.829 & 0.115 & 0.712 & 0.795 & 0.737 & 0.127 & 0.757 & 0.760 & 0.724 \\
		WS-SAM~\cite{he2023weaklysupervised}                & {\color[HTML]{00B0F0} \textbf{0.043}}& {0.721} & {0.839} & {0.816} & {0.037} & {0.694} & {0.849} & {0.797} & {0.046} & {0.878} & {\color[HTML]{00B0F0} \textbf{0.917}} & {0.877} & {0.078} &{0.858} &{\color[HTML]{00B0F0} \textbf{0.863}} & {0.775} & {0.089} & {0.839} & {0.841} & {\color[HTML]{00B0F0} \textbf{0.764}} \\
SEE~\cite{he2025segment} & {\color[HTML]{FF0000} \textbf{0.042}} & {\color[HTML]{00B0F0} \textbf{0.726}} & {\color[HTML]{00B0F0} \textbf{0.842}} & {\color[HTML]{00B0F0} \textbf{0.819}} & {\color[HTML]{FF0000} \textbf{0.035}} & {\color[HTML]{00B0F0} \textbf{0.705}} & {\color[HTML]{00B0F0} \textbf{0.857}} & {\color[HTML]{00B0F0} \textbf{0.800}} & {\color[HTML]{00B0F0} \textbf{0.045}} & {\color[HTML]{00B0F0} \textbf{0.883}} & {\color[HTML]{FF0000} \textbf{0.918}} & {\color[HTML]{00B0F0} \textbf{0.879}} & {\color[HTML]{00B0F0} \textbf{0.073}} & {\color[HTML]{00B0F0} \textbf{0.867}} & {\color[HTML]{FF0000} \textbf{0.871}} & {\color[HTML]{00B0F0} \textbf{0.777}} & {\color[HTML]{FF0000} \textbf{0.081}} & {\color[HTML]{00B0F0} \textbf{0.846}} & {\color[HTML]{00B0F0} \textbf{0.849}} & {\color[HTML]{FF0000} \textbf{0.768}} \\
\rowcolor{c2!20} SCALER (Ours) &{\color[HTML]{FF0000} \textbf{0.042}} & {\color[HTML]{FF0000} \textbf{0.731}} & {\color[HTML]{FF0000} \textbf{0.844}} & {\color[HTML]{FF0000} \textbf{0.822}} & {\color[HTML]{00B0F0} \textbf{0.036}} & {\color[HTML]{FF0000} \textbf{0.708}} & {\color[HTML]{FF0000} \textbf{0.861}} & {\color[HTML]{FF0000} \textbf{0.802}} & {\color[HTML]{FF0000} \textbf{0.044}} & {\color[HTML]{FF0000} \textbf{0.895}} & {\color[HTML]{00B0F0} \textbf{0.917}} & {\color[HTML]{FF0000} \textbf{0.883}} & {\color[HTML]{FF0000} \textbf{0.072}} & {\color[HTML]{FF0000} \textbf{0.871}} & {\color[HTML]{FF0000} \textbf{0.871}} & {\color[HTML]{FF0000} \textbf{0.779}} & {\color[HTML]{00B0F0} \textbf{0.083}} & {\color[HTML]{FF0000} \textbf{0.849}} & {\color[HTML]{FF0000} \textbf{0.853}} & {\color[HTML]{FF0000} \textbf{0.768}}   \\ 
  \bottomrule
	\end{tabular}} \label{table:MISTODWeak}
  \end{minipage} \\ \vspace{2mm}
  \begin{minipage}{\textwidth}
	\centering
	\setlength{\abovecaptionskip}{0cm}
	\caption{Results on PIS and TOD of the SSCOS task. 
	}
	\resizebox{\textwidth}{!}{
		\setlength{\tabcolsep}{0.8mm}
		\begin{tabular}{l|cccc|cccc|cccc|cccc|cccc}
			\toprule
			& \multicolumn{12}{c|}{Polyp Image Segmentation (PIS)}& \multicolumn{8}{c}{Transparent Object Detection (TOD)}\\ \cline{2-21}
			& \multicolumn{4}{c|}{\textit{CVC-ColonDB}}& \multicolumn{4}{c|}{\textit{ETIS}}& \multicolumn{4}{c|}{\textit{Kvasir}}& \multicolumn{4}{c|}{\textit{GDD}}& \multicolumn{4}{c}{\textit{GSD}}\\ \cline{2-21}
			\multirow{-3}{*}{Methods} & {\cellcolor{gray!40}$M$~$\downarrow$} &{\cellcolor{gray!40}$F_\beta$~$\uparrow$} &{\cellcolor{gray!40}$E_\phi$~$\uparrow$} & \multicolumn{1}{c|}{\cellcolor{gray!40}$S_\alpha$~$\uparrow$}& {\cellcolor{gray!40}$M$~$\downarrow$} &{\cellcolor{gray!40}$F_\beta$~$\uparrow$} &{\cellcolor{gray!40}$E_\phi$~$\uparrow$} & \multicolumn{1}{c|}{\cellcolor{gray!40}$S_\alpha$~$\uparrow$}& {\cellcolor{gray!40}$M$~$\downarrow$} &{\cellcolor{gray!40}$F_\beta$~$\uparrow$} &{\cellcolor{gray!40}$E_\phi$~$\uparrow$} & \multicolumn{1}{c|}{\cellcolor{gray!40}$S_\alpha$~$\uparrow$}& {\cellcolor{gray!40}$M$~$\downarrow$} &{\cellcolor{gray!40}$F_\beta$~$\uparrow$} &{\cellcolor{gray!40}$E_\phi$~$\uparrow$} & \multicolumn{1}{c|}{\cellcolor{gray!40}$S_\alpha$~$\uparrow$}& {\cellcolor{gray!40}$M$~$\downarrow$} &{\cellcolor{gray!40}$F_\beta$~$\uparrow$} &{\cellcolor{gray!40}$E_\phi$~$\uparrow$} & \multicolumn{1}{c}{\cellcolor{gray!40}$S_\alpha$~$\uparrow$}\\ \midrule
			\multicolumn{21}{c}{1/8 Labeled Training Data} \\ \midrule
			SAM~\cite{kirillov2023segment}        & 0.479                                 & 0.343                                 & 0.419                                 & 0.427                                 & 0.429                                 & 0.439                                 & 0.512                                 & 0.503                                 & 0.320                                 & 0.545                                 & 0.564                                 & 0.582                                 & 0.245                                 & 0.512                                 & 0.530                                 & 0.551                                 & 0.266                                 & 0.473                                 & 0.501                                 & 0.514                                 \\
SAM-S~\cite{chen2023sam}       & 0.185                                 & 0.517                                 & 0.627                                 & 0.661                                 & 0.172                                 & 0.558                                 & 0.706                                 & 0.682                                 & 0.131                                 & 0.738                                 & 0.772                                 & 0.783                                 & 0.185                                 & 0.674                                 & 0.703                                 & 0.618                                 & 0.177                                 & 0.688                                 & 0.724                                 & 0.652                                 \\
DTEN~\cite{ren2023towards}       & 0.073                                 & 0.607                                 & 0.722                                 & 0.743                                 & 0.064                                 & 0.607                                 & 0.759                                 & 0.757                                 & 0.078                                 & 0.837                                 & 0.867                                 & 0.833                                 & 0.096                                 & 0.815                                 & 0.818                                 & 0.733                                 & 0.102                                 & 0.803                                 & 0.792                                 & 0.736                                 \\
PGCL~\cite{basak2023pseudo}       & 0.068                                 & 0.629                                 & 0.743                                 & 0.749                                 & 0.057                                 & 0.623                                 & 0.771                                 & 0.764                                 & 0.070                                 & 0.838                                 & 0.869                                 & 0.855                                 & 0.097                                 & 0.819                                 & 0.823                                 & 0.747                                 & 0.095                                 & 0.817                                 & 0.804                                 & 0.741                                 \\
EPS~\cite{lee2023saliency}            & 0.060                                 & 0.645                                 & 0.758                                 & 0.756                                 & 0.053                                 & 0.642                                 & 0.786                                 & 0.772                                 & 0.063                                 & 0.847                                 & 0.883                                 & 0.858                                 & 0.085                                 & 0.837                                 & 0.842                                 & 0.759                                 & 0.089                                 & 0.829                                 & 0.817                                 & 0.754                                 \\
CoSOD~\cite{chakraborty2024unsupervised}      & 0.057                                 & 0.657                                 & 0.776                                 & 0.763                                 & 0.047                                 & 0.639                                 & 0.802                                 & 0.780                                 & 0.056                                 & 0.859                                 & 0.886                                 & 0.863                                 & 0.088                                 & 0.831                                 & 0.835                                 & 0.756                                 & 0.091                                 & 0.826                                 & 0.813                                 & 0.749                                 \\          
SEE~\cite{he2025segment} & {\color[HTML]{00B0F0} \textbf{0.045}} & {\color[HTML]{00B0F0} \textbf{0.706}} & {\color[HTML]{00B0F0} \textbf{0.828}} & {\color[HTML]{00B0F0} \textbf{0.801}} & {\color[HTML]{00B0F0} \textbf{0.039}} & {\color[HTML]{FF0000} \textbf{0.692}} & {\color[HTML]{00B0F0} \textbf{0.832}} & {\color[HTML]{00B0F0} \textbf{0.792}} & {\color[HTML]{00B0F0} \textbf{0.051}} & {\color[HTML]{00B0F0} \textbf{0.867}} & {\color[HTML]{00B0F0} \textbf{0.905}} & {\color[HTML]{00B0F0} \textbf{0.869}} & {\color[HTML]{00B0F0} \textbf{0.076}} & {\color[HTML]{00B0F0} \textbf{0.846}} & {\color[HTML]{FF0000} \textbf{0.855}} & {\color[HTML]{00B0F0} \textbf{0.768}} & {\color[HTML]{00B0F0} \textbf{0.085}} & {\color[HTML]{00B0F0} \textbf{0.835}} & {\color[HTML]{00B0F0} \textbf{0.832}} & {\color[HTML]{00B0F0} \textbf{0.763}} \\ 
\rowcolor{c2!20} SCALER (Ours) &{\color[HTML]{FF0000} \textbf{0.043}} & {\color[HTML]{FF0000} \textbf{0.726}} & {\color[HTML]{FF0000} \textbf{0.841}} & {\color[HTML]{FF0000} \textbf{0.823}} & {\color[HTML]{FF0000} \textbf{0.038}} & {\color[HTML]{00B0F0} \textbf{0.685}} & {\color[HTML]{FF0000} \textbf{0.851}} & {\color[HTML]{FF0000} \textbf{0.798}} & {\color[HTML]{FF0000} \textbf{0.050}} & {\color[HTML]{FF0000} \textbf{0.873}} & {\color[HTML]{FF0000} \textbf{0.914}} & {\color[HTML]{FF0000} \textbf{0.879}} & {\color[HTML]{FF0000} \textbf{0.073}} & {\color[HTML]{FF0000} \textbf{0.854}} & {\color[HTML]{00B0F0} \textbf{0.849}} & {\color[HTML]{FF0000} \textbf{0.781}} & {\color[HTML]{FF0000} \textbf{0.083}} & {\color[HTML]{FF0000} \textbf{0.845}} & {\color[HTML]{FF0000} \textbf{0.841}} & {\color[HTML]{FF0000} \textbf{0.777}} \\ \midrule
\multicolumn{21}{c}{1/16 Labeled Training Data} \\ \midrule
SAM~\cite{kirillov2023segment}        & 0.479                                 & 0.343                                 & 0.419                                 & 0.427                                 & 0.429                                 & 0.439                                 & 0.512                                 & 0.503                                 & 0.320                                 & 0.545                                 & 0.564                                 & 0.582                                 & 0.245                                 & 0.512                                 & 0.530                                 & 0.551                                 & 0.266                                 & 0.473                                 & 0.501                                 & 0.514                                 \\
SAM-S~\cite{chen2023sam}        & 0.201                                 & 0.471                                 & 0.593                                 & 0.643                                 & 0.183                                 & 0.533                                 & 0.677                                 & 0.660                                 & 0.145                                 & 0.712                                 & 0.753                                 & 0.762                                 & 0.203                                 & 0.657                                 & 0.672                                 & 0.604                                 & 0.193                                 & 0.654                                 & 0.684                                 & 0.627                                 \\
DTEN~\cite{ren2023towards}      & 0.080                                 & 0.586                                 & 0.696                                 & 0.732                                 & 0.071                                 & 0.592                                 & 0.730                                 & 0.750                                 & 0.086                                 & 0.813                                 & 0.824                                 & 0.833                                 & 0.112                                 & 0.797                                 & 0.810                                 & 0.736                                 & 0.115                                 & 0.786                                 & 0.770                                 & 0.705                                 \\
PGCL~\cite{basak2023pseudo}     & 0.075                                 & 0.608                                 & 0.719                                 & 0.745                                 & 0.063                                 & 0.614                                 & 0.753                                 & 0.754                                 & 0.078                                 & 0.836                                 & 0.845                                 & 0.835                                 & 0.108                                 & 0.794                                 & 0.803                                 & 0.735                                 & 0.108                                 & 0.800                                 & 0.788                                 & 0.718                                 \\
EPS~\cite{lee2023saliency}     & 0.065                                 & 0.631                                 & 0.746                                 & 0.750                                 & 0.057                                 & 0.631                                 & 0.774                                 & 0.755                                 & 0.068                                 & 0.835                                 & 0.863                                 & 0.846                                 & {0.089}                                & {0.818}                                 & {0.825}                                 & {0.750}                                 & {0.097}                                 & {0.812}                                 & {0.803}                                 & {0.737}                                 \\
CoSOD~\cite{chakraborty2024unsupervised}     & {0.063}                                 & {0.638}                                 & {0.754}                                 & {0.755}                                 & {0.050}                                 & {0.632}                                 & {0.781}                                 & {0.767}                                 & {0.066}                                 & {0.840}                                 & {0.871}                                 & {0.852}                                 & 0.095                                 & 0.813                                 & 0.816                                 & 0.746                                 & 0.105                                 & 0.793                                 & 0.792                                 & 0.732                                 \\
\rowcolor{c2!20} SEE (Ours)    & {\color[HTML]{00B0F0} \textbf{0.050}} & {\color[HTML]{00B0F0} \textbf{0.693}} & {\color[HTML]{00B0F0} \textbf{0.812}} & {\color[HTML]{FF0000} \textbf{0.795}} & {\color[HTML]{00B0F0} \textbf{0.043}} & {\color[HTML]{00B0F0} \textbf{0.674}} & {\color[HTML]{00B0F0} \textbf{0.811}} & {\color[HTML]{00B0F0} \textbf{0.784}} & {\color[HTML]{00B0F0} \textbf{0.057}} & {\color[HTML]{00B0F0} \textbf{0.858}} & {\color[HTML]{00B0F0} \textbf{0.893}} & {\color[HTML]{00B0F0} \textbf{0.861}} & {\color[HTML]{00B0F0} \textbf{0.080}} & {\color[HTML]{00B0F0} \textbf{0.827}} & {\color[HTML]{00B0F0} \textbf{0.831}} & {\color[HTML]{00B0F0} \textbf{0.759}} & {\color[HTML]{00B0F0} \textbf{0.092}} & {\color[HTML]{FF0000} \textbf{0.824}} & {\color[HTML]{00B0F0} \textbf{0.817}} & {\color[HTML]{00B0F0} \textbf{0.750}} \\
\rowcolor{c2!20} SCALER (Ours) & {\color[HTML]{FF0000} \textbf{0.048}} & {\color[HTML]{FF0000} \textbf{0.698}} & {\color[HTML]{FF0000} \textbf{0.821}} & {\color[HTML]{00B0F0} \textbf{0.777}} & {\color[HTML]{FF0000} \textbf{0.041}} & {\color[HTML]{FF0000} \textbf{0.686}} & {\color[HTML]{FF0000} \textbf{0.822}} & {\color[HTML]{FF0000} \textbf{0.799}} & {\color[HTML]{FF0000} \textbf{0.056}} & {\color[HTML]{FF0000} \textbf{0.878}} & {\color[HTML]{FF0000} \textbf{0.899}} & {\color[HTML]{FF0000} \textbf{0.885}} & {\color[HTML]{FF0000} \textbf{0.078}} & {\color[HTML]{FF0000} \textbf{0.851}} & {\color[HTML]{FF0000} \textbf{0.848}} & {\color[HTML]{FF0000} \textbf{0.767}} & {\color[HTML]{FF0000} \textbf{0.082}} & {\color[HTML]{00B0F0} \textbf{0.819}} & {\color[HTML]{FF0000} \textbf{0.831}} & {\color[HTML]{FF0000} \textbf{0.768}} \\
  \bottomrule
	\end{tabular}} \label{table:MISTODSemi}
	\vspace{-0.8mm}
  \end{minipage}

\end{table*}

\subsection{Optimization Procedure}
The full SCALER pipeline proceeds in three stages to ensure stable convergence toward mutual enhancement.

\textbf{(1) Stage 1 (Initialization)}: We first pretrain the mean-teacher framework and fine-tune SAM separately using only the available labeled COS data. This provides both models with a strong, task-specific starting point.

\textbf{(2) Stage 2 (Segmenter Warm-up)}: We fix the initialized SAM and run only Phase \uppercase\expandafter{\romannumeral1} on the full label-deficient dataset. This allows the segmenter to quickly develop strong specialization by leveraging SAM's generalist pseudo-labels.

\textbf{(3) Stage 3 (Iterative Mutual Enhancement)}: This is the core. We alternately optimize Phase \uppercase\expandafter{\romannumeral1} (enhancing the segmenter with SAM) and Phase \uppercase\expandafter{\romannumeral2} (enhancing SAM via the segmenter). Each model generates pseudo-labels that fit their corresponding characteristic for the other, facilitating the iterative mutual enhancement described in our contributions.

\section{Experiments} \label{Sec:Experiments}
\subsection{Experimental Setup}
\noindent \textbf{Implementation details}. We implement our method in PyTorch using two RTX 6000 Ada GPUs. Following~\cite{he2023weaklysupervised}, we adopt ResNet50 pre-trained on ImageNet~\cite{deng2009imagenet} as the default backbone. All input images are resized to $352 \times 352$ for both training and testing. The Adam optimizer is used for training, with momentum parameters set to $(0.9, 0.999)$. The batch size is fixed at 12, and the initial learning rate is set to 0.0001, which is reduced by a factor of 0.1 every 80 epochs. The number of augmentation images $K$ is set to 12. In this paper, our segmenter is the same as WS-SAM~\cite{he2023weaklysupervised} for fairness. We fine-tune SAM using SAM-Adapter~\cite{chen2023sam}, but other adaptation strategies and foundation models (\textit{e.g.}, SAM2) can be seamlessly integrated into SCALER. In general, stronger foundation models and/or more effective fine-tuning strategies provide higher-quality pseudo-labels, yielding further gains for the segmenter and, via Phase \uppercase\expandafter{\romannumeral2}, additional improvements for the foundation model itself.

\subsection{Comparative Evaluation} \vspace{-1mm}
Following~\cite{he2023weaklysupervised}, we report results with point annotations under weak supervision. For the COD task, we further assess performance with scribble annotations provided by~\cite{gao2022weakly}.

\noindent \textbf{Camouflaged object detection}. Tables \ref{table:CODWeak} and \ref{table:CODSemi} show that our SCALER framework outperforms competing methods in both weakly supervised and semi-supervised settings. Although fine-tuning SAM via SAMadapter~\cite{chen2023sam} under point ($\text{SAM-W}_\text{P}$), scribble ($\text{SAM-W}_\text{S}$), and semi-supervised configurations yields performance improvements, these variants still underperform relative to SCALER. Furthermore, SCALER surpasses the previously proposed WS-SAM~\cite{he2023weaklysupervised} and SEE~\cite{he2025segment} frameworks across all evaluation protocols, demonstrating its effectiveness in producing higher-quality pseudo-labels. Fig. \ref{fig:COSQuali} provides qualitative comparisons under point supervision, further illustrating the advantages of our alternating optimization strategy.

\noindent \textbf{Polyp image segmentation}. Polyps often exhibit strong visual similarity to surrounding tissue, making polyp image segmentation (PIS) particularly challenging. As shown in Tables~\ref{table:MISTODWeak} and~\ref{table:MISTODSemi}, our SCALER framework significantly outperforms the next best method, SEE, under point supervision. Both SAM and $\text{SAM-W}_\text{P}$ perform poorly on this task, underscoring their limitations in handling concealed polyps. Fig. \ref{fig:COSQuali} illustrates that our approach more accurately localizes hidden polyps and produces segmentation boundaries that closely match the true lesion contours.

\noindent \textbf{Transparent object detection}. TOD is essential for enhancing vision systems in robotics, yet it remains challenging due to the minimal visual contrast of objects. As shown in Tables~\ref{table:MISTODWeak} and~\ref{table:MISTODSemi}, our SCALER framework surpasses current state-of-the-art methods under both weakly and semi-supervised settings. Fig.~\ref{fig:COSQuali} illustrates that our method not only localizes transparent objects with high accuracy but also refines their segmentation boundaries and effectively suppresses background clutter.
\begin{table*}[t]
 \begin{minipage}{\textwidth}
 \centering
	\setlength{\abovecaptionskip}{0cm}
	\caption{Ablation study in the COD task on the \textit{COD10K} dataset with the scribble supervision.
	}\label{table:ablation}
\resizebox{\textwidth}{!}{
		\setlength{\tabcolsep}{0.5mm}
\begin{tabular}{l|ccccc|ccccc|cc|c}
\toprule
\multirow{2}{*}{Metrics} & \multicolumn{5}{c|}{Phase I}                                                        & \multicolumn{5}{c|}{Phase II}                                                                            & \multicolumn{2}{c|}{Optimization} & \cellcolor{c2!20}SCALER \\ \cline{2-13}
& w/o $PL_F$ & w/o $E(\bigcdot)$ & w/o $U(\bigcdot)$ & $R_1(\bigcdot)\!\rightarrow\! R(\bigcdot)$ & $R_2(\bigcdot)\!\rightarrow\! R(\bigcdot)$ & w/o Phase II & w/o $\mathcal{L}_{ai}$ & $\mathcal{L}_{ai}^1 \rightarrow \mathcal{L}_{ai}$ & w/o $\mathcal{L}_{nr}$ & $\mathcal{L}_{nr}^1 \rightarrow \mathcal{L}_{nr}$ & w/o step 1    & w/o step 2   &\cellcolor{c2!20} (Ours)   \\ \midrule
$M$~$\downarrow$                       & 0.039           & 0.036         & 0.038         & 0.037                       & 0.035            &   0.037        & 0.036             & 0.039                                    &0.036              & 0.035                                    & 0.040              &0.036              & \cellcolor{c2!20} 0.034  \\
$F_\beta$~$\uparrow$                        &0.732            &0.724          &0.721          &0.728                        &0.722            &   0.723         &0.733              &0.730                                     &0.729              &0.734                                     &0.720               &0.733              & \cellcolor{c2!20} 0.736  \\
$E_\phi$~$\uparrow$ &0.863            &0.870          &0.869          &0.866                        &0.868        &       0.860         &0.865              &0.869                                     &0.870              &0.879                                     &0.864               &0.875              & \cellcolor{c2!20} 0.882  \\
$S_\alpha$~$\uparrow$ &0.804            &0.805          &0.813          &0.806                        &0.808          &         0.805      &0.812              &0.804                                     &0.807              &0.809                                     &0.802               &0.811              & \cellcolor{c2!20} 0.815  \\ \bottomrule
\end{tabular}}
  \end{minipage}\\
 \begin{minipage}{0.436\textwidth}
    \centering
	\setlength{\abovecaptionskip}{0cm}
	\caption{Generalization of SCALER. ``+'' means integrating the method with our SCALER framework. Our SCALER serves as a plug-and-play framework for existing methods.
	}\label{table:generalization}
\resizebox{\textwidth}{!}{
		\setlength{\tabcolsep}{0.6mm}
\begin{tabular}{l|cc|cc|cc}
\toprule

Metrics& GenSAM & \cellcolor{c2!20} GenSAM+ & WS-SAM & \cellcolor{c2!20} WS-SAM+ & SEE   & \cellcolor{c2!20} SEE+ \\ \midrule
$M$~$\downarrow$       & 0.067  & \cellcolor{c2!20} 0.061   & 0.038  & \cellcolor{c2!20} 0.036   & 0.036 & \cellcolor{c2!20} 0.035  \\
$F_\beta$~$\uparrow$ & 0.681  & \cellcolor{c2!20} 0.693   & 0.719  & \cellcolor{c2!20} 0.728   & 0.729 & \cellcolor{c2!20} 0.738   \\
$E_\phi$~$\uparrow$ & 0.838  & \cellcolor{c2!20} 0.851   & 0.878  & \cellcolor{c2!20} 0.887   & 0.883 & \cellcolor{c2!20} 0.886   \\
$S_\alpha$~$\uparrow$ & 0.775  & \cellcolor{c2!20} 0.780   & 0.803  & \cellcolor{c2!20} 0.809   & 0.807 & \cellcolor{c2!20} 0.813 \\ \bottomrule  
\end{tabular}} 
\end{minipage}
 \begin{minipage}{0.554\textwidth}
    \centering
	\setlength{\abovecaptionskip}{0cm}
	\caption{Performance of foundation models with different training manners, ``SAM'' and ``SAM2'' are trained under our SCALER framework, while the two ``adapters'' indicate direct fine-tuning of the foundation models.
	}\label{table:generalization}
\resizebox{\textwidth}{!}{
		\setlength{\tabcolsep}{0.6mm}
\begin{tabular}{l|ccc|ccc}
\toprule
Metrics& \cellcolor{c2!20} SCALER+S & \cellcolor{c2!20} SAM   & SAMadapter~\cite{chen2023sam} & \cellcolor{c2!20} SCALER+S2 & \cellcolor{c2!20} SAM2  & SAM2adapter \cite{chen2024sam2} \\ \midrule
$M$~$\downarrow$       & \cellcolor{c2!20} 0.034    & \cellcolor{c2!20} 0.027 & 0.036      & \cellcolor{c2!20} 0.033     & \cellcolor{c2!20} 0.023 & 0.035       \\
$F_\beta$~$\uparrow$ & \cellcolor{c2!20} 0.736    & \cellcolor{c2!20} 0.765 & 0.732      & \cellcolor{c2!20} 0.741     & \cellcolor{c2!20} 0.778 & 0.733       \\
$E_\phi$~$\uparrow$ & \cellcolor{c2!20} 0.882    & \cellcolor{c2!20} 0.912 & 0.888      & \cellcolor{c2!20} 0.889     & \cellcolor{c2!20} 0.920 & 0.886       \\
$S_\alpha$~$\uparrow$ & \cellcolor{c2!20} 0.815    & \cellcolor{c2!20} 0.831 & 0.818      & \cellcolor{c2!20} 0.819     & \cellcolor{c2!20} 0.835 & 0.820  \\ \bottomrule  
\end{tabular}}
\end{minipage}\vspace{-7mm}
\end{table*}

\subsection{Ablation Study}\label{ablation-study}
Experiments are conducted on the \textit{COD10K} dataset with scribble supervision in~\cref{ablation-study,Analysis}.

\noindent \textbf{Effectiveness in Phase I}. As shown in~\cref{table:ablation}, in Phase I, we observe performance degradation upon removing the fused pseudo-label $PL_F$, the entropy-based image-level weighting mechanism $E(\cdot)$, or the uncertainty-based pixel-level weighting mechanism $U(\cdot)$. Besides, omitting our specialized refinement procedures for difficult samples (denoted $R_1(\cdot)$) and simple samples (denoted $R_2(\cdot)$) also leads to a noticeable performance drop. These results demonstrate the effectiveness of our strategies.

\noindent \textbf{Effectiveness in Phase II}. As shown in~\cref{table:ablation}, removing Phase II entirely (\textit{i.e.}, running only Stage 1 and 2) reverts SCALER to a one-way framework. This ``w/o Phase II'' ablation shows a significant performance drop, confirming that the mutual enhancement from Phase II is critical. We further evaluate the proposed augmentation invariance loss $L_{ai}$ and noise resistance loss $L_{nr}$. Specifically, using an alternative $L_{ai}^1$, which imposes invariance between two weak augmentations only, or employing $L_{nr}^1$, which directly applies the refinement strategy $R(\cdot)$ from Phase I, results in decreased performance. This confirms that our proposed loss functions are better tailored for effectively fine-tuning SAM.

\noindent \textbf{Impact of the optimization procedure}. To enhance the capability of both SAM and the mean-teacher framework in generating high-quality pseudo-labels, we structure our optimization procedure into three sequential steps, with the initial two steps dedicated to network pretraining. As indicated in~\cref{table:ablation}, each of these pretraining steps contributes positively to the overall segmentation performance, underscoring their importance in our proposed method.

\vspace{-2mm}
\subsection{Further Analysis}\label{Analysis}\vspace{-2mm}
Results on salient object detection are placed in the Supp. 

\noindent \textbf{Generalizability of our SCALER}. 
As shown in~\cref{table:generalization}, integrating SCALER with cutting-edge methods (indicated by the suffix “+”) yields performance gains. This demonstrates the potential of SCALER to function as a plug-and-play framework for enhancing existing segmentation models.

\noindent \textbf{Performance of foundation models with different training manners}. As shown in~\cref{table:generalization}, foundation models trained within our SCALER framework achieve substantially better performance than those directly fine-tuned with adapter-based methods. We also report results of our segmenter using SAM and SAM2 as the foundation models, denoted as ``+S'' and ``+S2'' respectively. Notably, our segmenter even surpasses the large foundation models fine-tuned via adapters, demonstrating the effectiveness and superiority of the proposed bi-directional collaborative learning framework.

\vspace{-2mm}
\section{Discussions}\vspace{-2mm}
Beyond proposing a robust LDCOS framework, SCALER reveals an important insight from its bi-directional design: a lightweight, task-specific segmenter can effectively guide a large, generalist foundation model—such as SAM—under label-deficient conditions. This finding underscores the advantage of mutual knowledge transfer over the one-way distillation used in previous works\cite{he2023weaklysupervised,he2025segment}. By enabling the foundation model to benefit from the segmenter’s domain specialization, SCALER improves adaptation in challenging settings (e.g., medical imaging)\cite{ji2023sam,ji2023segment,mazurowski2023segment}, suggesting a broader paradigm where knowledge is exchanged reciprocally rather than unidirectionally.

\vspace{-2mm}
\section{Limitations and Future Work} \label{Sec:Limitation}\vspace{-2mm}
SCALER still has two main limitations.
First, when both the segmenter and SAM simultaneously fail to localize an object, the mutual learning loop cannot start effectively. Future work will explore introducing auxiliary knowledge such as text prompts or generative priors to provide initial guidance and overcome this cold-start issue.
Second, the proposed bi-directional learning paradigm, where a lightweight, task-specific model and a large foundation model mutually enhance each other, has broader potential. We plan to extend this principle to other label-deficient vision tasks, including object detection and depth estimation.

\vspace{-2mm}
\section{Conclusions}\vspace{-2mm}
\label{Sec:Conclusion}
We propose SCALER, a unified framework for LDCOS, combining a mean-teacher approach for consistency regularization with a learnable SAM-based model for knowledge distillation. In Phase \uppercase\expandafter{\romannumeral1}, with SAM fixed, we refine the segmenter using image-level and pixel-level weighting schemes that select trustworthy regions in pseudo-labels. In Phase \uppercase\expandafter{\romannumeral2}, with the segmenter fixed, we update SAM by introducing an augmentation invariance loss and a noise resistance loss, promoting stable representation learning.

\newpage
{
    \small
    \bibliographystyle{ieeenat_fullname}
    \bibliography{main}
}

\end{document}